\documentclass[10pt,twocolumn,letterpaper]{article}

\usepackage{iccv}
\usepackage{times}
\usepackage{epsfig}
\usepackage{graphicx}
\usepackage{amsmath}
\usepackage{amssymb}
\usepackage{algorithm}
\usepackage{algorithmic}
\usepackage{gensymb}
\usepackage{float}

\usepackage{multirow}
\usepackage{multicol}

\usepackage[pagebackref=true,breaklinks=true,letterpaper=true,colorlinks,bookmarks=false]{hyperref}

\iccvfinalcopy 

\def\iccvPaperID{****} 

\ificcvfinal\fi

\begin{document}

\title{End-to-End Semi-Supervised Object Detection with Soft Teacher}

\author{
Mengde Xu$^{1 \dag}$\thanks{Equal contribution. \dag This work is done when Mengde Xu was intern in MSRA. \ddag Contact person.}
\quad Zheng~Zhang$^{1,2}$\footnotemark[1]~~\footnotemark[3]
\quad Han~Hu$^2$\footnotemark[3]
\quad Jianfeng Wang$^2$
\quad Lijuan~Wang$^2$ 
\quad Fangyun~Wei$^2$ \\
\quad Xiang~Bai$^1$
\quad Zicheng Liu$^2$
\\
{$^1$Huazhong University of Science and Technology} \\
\small{\texttt{\{mdxu,xbai\}@hust.edu.cn}} \\
{$^2$Microsoft} \\
\small{
\texttt{\{zhez,hanhu,jianfw,lijuanw,fawe,zliu\}@microsoft.com}}
}
\maketitle
\ificcvfinal\thispagestyle{empty}\fi

\begin{abstract}
This paper presents an end-to-end semi-supervised object detection approach, in contrast to previous more complex multi-stage methods. The end-to-end training gradually improves pseudo label qualities during the curriculum, and the more and more accurate pseudo labels in turn benefit object detection training. We also propose two simple yet effective techniques within this framework: a soft teacher mechanism where the classification loss of each unlabeled bounding box is weighed by the classification score produced by the teacher network; a box jittering approach to select reliable pseudo boxes for the learning of box regression. On the COCO benchmark, the proposed approach outperforms previous methods by a large margin under various labeling ratios, i.e. 1\%, 5\% and 10\%. Moreover, our approach proves to perform also well when the amount of labeled data is relatively large. For example, it can improve a 40.9 mAP baseline detector trained using the full COCO training set by +3.6 mAP, reaching 44.5 mAP, by leveraging the 123K unlabeled images of COCO. On the state-of-the-art Swin Transformer based object detector (58.9 mAP on test-dev), it can still significantly improve the detection accuracy by +1.5 mAP, reaching 60.4 mAP, and improve the instance segmentation accuracy by +1.2 mAP, reaching 52.4 mAP. Further incorporating with the Object365 pre-trained model, the detection accuracy reaches 61.3 mAP and the instance segmentation accuracy reaches 53.0 mAP, pushing the new state-of-the-art. The code and models will be made publicly available at~\url{https://github.com/microsoft/SoftTeacher}.
\end{abstract}

\vspace{-1em}
\section{Introduction}
\begin{table*}
\small
\begin{center}
\begin{tabular}{c|c|c|c|c|c}
\hline
\multirow{2}{*}{detector} & \multirow{2}{*}{method} & \multicolumn{2}{c}{val2017} & \multicolumn{2}{c}{test-dev2017} \\
& & $\text{mAP}^{\text{det}}$ & $\text{mAP}^{\text{mask}}$ & $\text{mAP}^{\text{det}}$ & $\text{mAP}^{\text{mask}}$  \\
\hline
\multirow{3}{*}{HTC++(Swin-L) w/ single-scale} & supervised & 57.1 & 49.6 & - & - \\
& ours & 59.1 & 51.0 & - & - \\
& ours$^{*}$ & 60.1 & 51.9 & - & - \\
\hline
\multirow{3}{*}{HTC++(Swin-L) w/ multi-scale} & supervised& 58.2 & 50.5 & 58.9 & 51.2 \\
& ours & 59.9 & 51.9 & 60.4  & 52.4  \\
& ours$^{*}$ & 60.7  & 52.5 & 61.3 & 53.0  \\
\end{tabular}
\end{center}
\caption{On the state-of-the-art detector HTC++(Swin-L), our method surpasses the supervised learning on both \texttt{val2017} and \texttt{test-dev2017}. * indicates that models are pre-trained with Object365~\cite{shao2019objects365} dataset.}
\label{tab::teaser_table}
\end{table*}

Data matters. In fact, large data such as ImageNet has largely triggered the boom of deep learning in computer vision. However, obtaining labels can be a bottleneck, due to the time-consuming and expensive annotation process. This has encouraged learning methods to leverage unlabeled data in training deep neural models, such as self-supervised learning and semi-supervised learning. This paper studies the problem of semi-supervised learning, in particular for object detection.

For semi-supervised object detection, we are concerned with the pseudo-label based approaches, which are the current state-of-the-art. These approaches~\cite{sohn2020simple,zoph2020rethinking} conduct a multi-stage training schema, with the first stage training an initial detector using labeled data, followed by a pseudo-labeling process for unlabeled data and a re-training step based on the pseudo labeled unannotated data. These multi-stage approaches achieve reasonably good accuracy, however, the final performance is limited by the quality of pseudo labels generated by an initial and probably inaccurate detector trained using a small amount of labeled data.

\begin{figure}
    \centering
    \includegraphics[width=1.0\columnwidth]{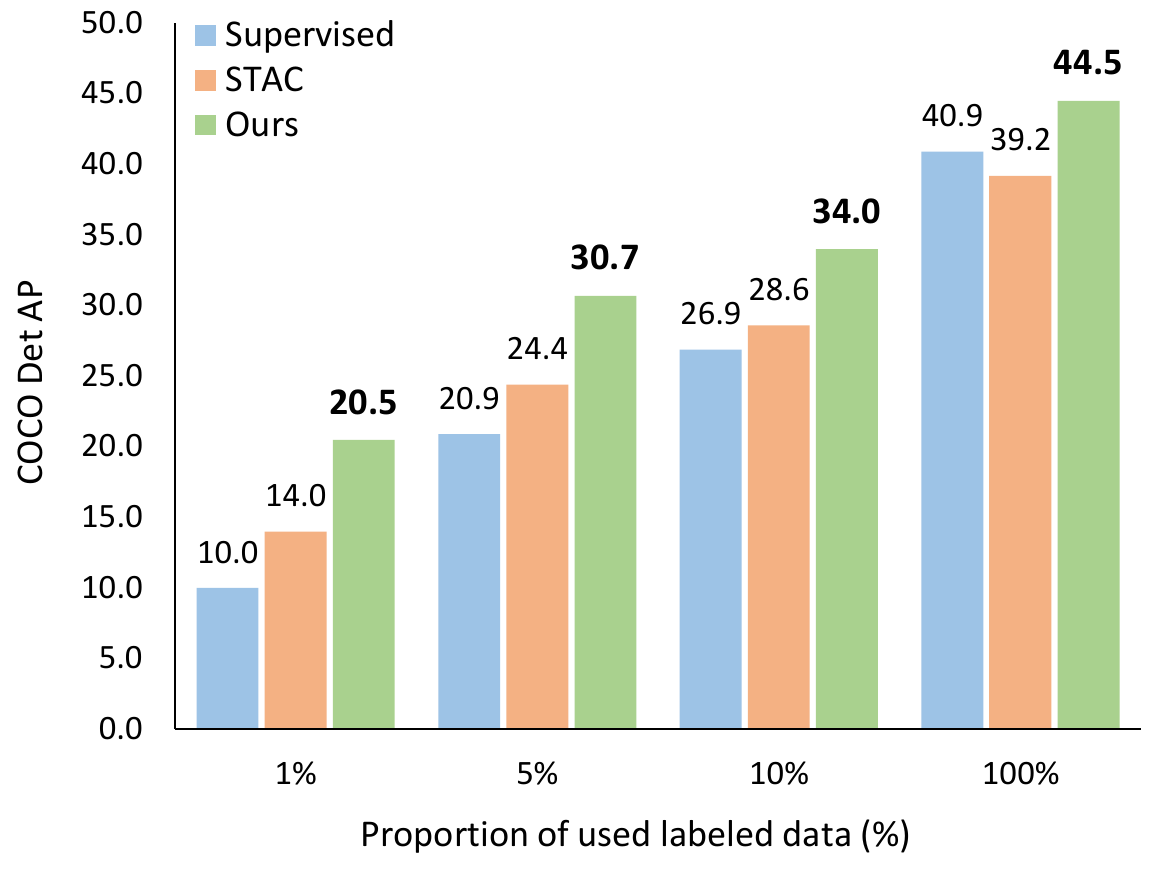}
    \caption{The proposed end-to-end pseudo-label based semi-supervised object detection method outperforms the STAC~\cite{sohn2020simple} by a large margin on MS-COCO benchmark.}
    \label{fig:teaser}
\end{figure}

To address this issue, we present an end-to-end pseudo-label based semi-supervised object detection framework, which simultaneously  performs pseudo-labeling for unlabeled images and trains a detector using these pseudo labels along with a few labeled ones at each iteration. Specifically, labeled and unlabeled images are randomly sampled with a preset ratio to form one data batch. Two models are applied on these images, with one conducting detection training and the other in charge of annotating pseudo labels for unlabeled images. The former is also referred to as a student, and the latter is a teacher, which is an exponential moving average (EMA) of the student model. This end-to-end approach avoids the complicated multi-stage training scheme. Moreover, it also enables a ``flywheel effect'' that the pseudo labeling and the detection training processes can mutually reinforce each other, so that both get better and better as the training goes on.

Another important benefit of this end-to-end framework is that it allows for greater leverage of the teacher model to guide the training of the student model, rather than just providing ``some generated pseudo boxes with hard category labels'' as in previous approaches~\cite{sohn2020simple,zoph2020rethinking}. 
\begin{figure*}
    \centering
    \includegraphics[width=1.0\linewidth]{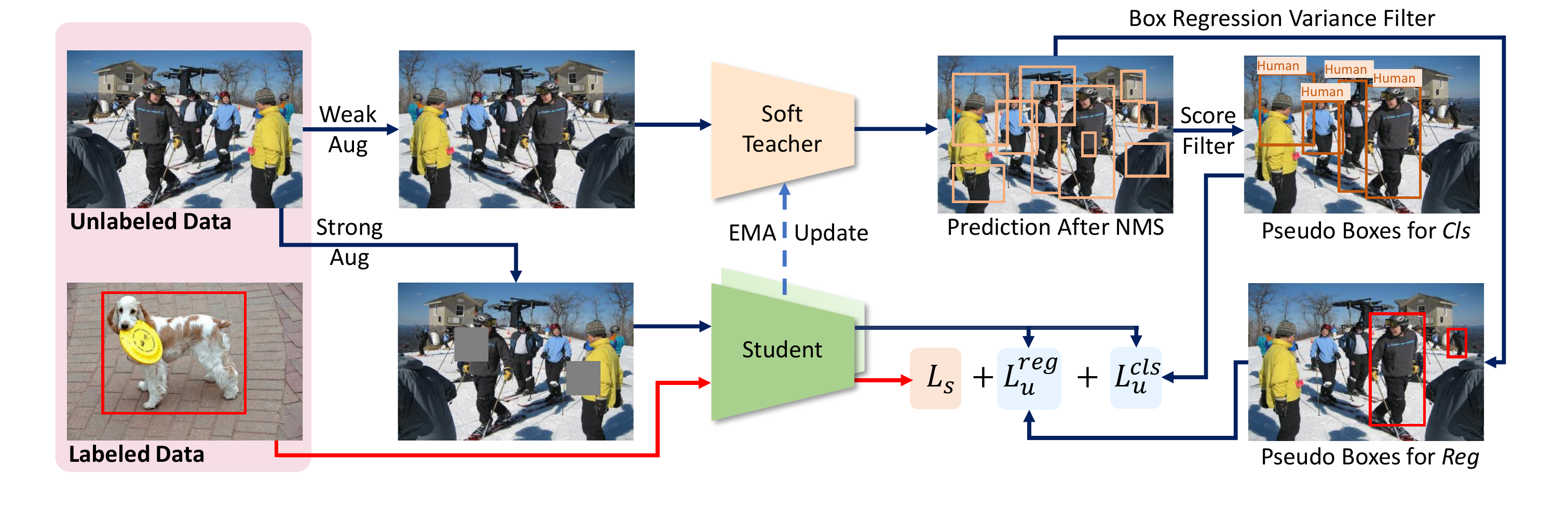}
    \caption{The overview of the end-to-end pseudo-labeling framework for semi-supervised object detection. Unlabeled images and labeled images form the training data batch. In each training iteration, a soft teacher is applied to perform pseudo-labeling on weak augmented unlabeled images on the fly. Two sets of pseudo boxes are produced: one is used for classification branch by filtering boxes according to the foreground score, and the other is used for box regression branch by filtering boxes according to box regression variance. The teacher model is updated by student model via exponential mean average (EMA) manner. The final loss is the sum of supervised detection loss $L_s$ and unsupervised detection loss $L_u$.}
    \label{fig:pipeline}
\end{figure*} A \emph{soft teacher} approach is proposed to implement this insight. In this approach, the teacher model is used to directly assess all the box candidates that are generated by the student model, rather than providing ``pseudo boxes'' to assign category labels and regression vectors to these student-generated box candidates. The direct assessment on these box candidates enables more extensive supervision information to be used in the student model training. Specifically, we first categorize the box candidates as foreground/background by their detection scores with a high foreground threshold to ensure a high precision of the positive pseudo labels, as in~\cite{sohn2020simple}. This high foreground threshold, however, results in many positive box candidates mistakenly assigned as background. To address this issue, we propose using a \emph{reliability} measure to weight the loss of each ``background'' box candidate. We empirically find that a simple detection score produced by the teacher model can well serve as the reliability measure, and is used in our approach. We find that this approach measure performs significantly better than previous hard foreground/background assignment methods (see Table~\ref{tab::system_comparision_coco} and Table~\ref{tab::system_full_coco_others}), and we name it ``\emph{soft teacher}''.

Another approach instantiates this insight is to select reliable bounding boxes for the training of the student's localization branch, by a \emph{box jittering} approach. This approach first jitters a pseudo-foreground box candidate several times. Then these jittered boxes are regressed according the teacher model's location branch, and the variance of these regressed boxes is used as a \emph{reliability measure}. The box candidate with adequately high reliability will be used for the training of the student's localization branch.

On MS-COCO object detection benchmark~\cite{lin2014microsoft}, our approach achieves 20.5 mAP, 30.7 mAP and 34.0 mAP on \texttt{val2017} with 1\%, 5\% and 10\% labeled data using the Faster R-CNN~\cite{ren2015faster} framework with ResNet-50~\cite{he2016deep} and FPN~\cite{lin2017feature}, surpassing previous best method STAC~\cite{sohn2020simple} by \textbf{+6.5}, \textbf{+6.4} and \textbf{+5.4} mAP, respectively. 

In addition, we also perform evaluation on a more challenge setting where the labelled data has been adequately large to train a reasonably accurate object detector. Specifically, we adopt the complete COCO \texttt{train2017} set as labeled data and the \texttt{unlabeled2017} set as the unlabeled data. Under this setting, we improve the supervised baseline of a Faster R-CNN approach with ResNet-50 and ResNet-101 backbones by \textbf{+3.6} mAP and \textbf{+3.0} mAP, respectively.

Moreover, on a state-of-the-art Swin-Transformer~\cite{liu2021swin} based detector which achieves 58.9 mAP for object detection and 51.2 mAP for instance segmentation on COCO \texttt{test-dev2017}, the proposed approach can still improve the accuracy by +1.5 mAP and +1.2 mAP, respectively, reaching \textbf{60.4} mAP and \textbf{52.4} mAP. Further incorporating with the Object365~\cite{shao2019objects365} pre-trained model, the detection accuracy reaches \textbf{61.3} mAP and the instance segmentation accuracy reaches \textbf{53.0} mAP, which is the new state-of-the-art on this benchmark.

\section{Related works}
\paragraph{Semi-supervised learning in image classification}

Semi-supervised learning in image classification can be roughly categorized into two groups: consistency based and pseudo-label based. The consistency based methods~\cite{bachman2014learning,sajjadi2016regularization,miyato2018virtual,laine2016temporal} leverage the unlabeled images to construct a regularization loss which encourages different perturbations of a same image to produce similar predictions.
There are several ways to implement perturbations, including
perturbing the model~\cite{bachman2014learning}, augmenting the images~\cite{sajjadi2016regularization} or adversarial training~\cite{miyato2018virtual}. In~\cite{laine2016temporal}, the training target is assembled by predicting different training steps. In~\cite{tarvainen2017mean}, they develop~\cite{laine2016temporal} by ensembling the model itself instead of the model prediction, the so-called exponential mean average (EMA) of the student model.
The pseudo-label approaches~\cite{xie2020self,grandvalet2005semi,lee2013pseudo} (also named as self-training) annotate unlabeled images with pseudo labels by an initially trained classification model, and the detector is refined by these pseudo labeled images. Unlike our method which focuses on object detection, the pseudo-label 
does not have to solve the problem of assigning foreground/background labels and box regression when classifying images. Recently, some works~\cite{xie2019unsupervised,berthelot2019mixmatch,berthelot2019remixmatch,sohn2020fixmatch} explore the importance of data augmentation in semi-supervised learning, which inspire us to use the weak augmentation to generate pseudo-labels and the strong augmentation for the learning of detection models.

\paragraph{Semi-supervised learning in object detection}
Similar to the semi-supervised learning in image classification, semi-supervised object detection methods also have two categories: the consistency methods~\cite{jeong2019consistency, tang2021proposal} and pseudo-label methods~\cite{radosavovic2018data, zoph2020rethinking, li2020improving, sohn2020simple, wang2018towards}. Our method belongs to the pseudo-label category. In~\cite{radosavovic2018data, zoph2020rethinking}, the predictions of different data augmentation are ensembled to form the pseudo labels for unlabeled images. In~\cite{li2020improving}, a SelectiveNet is trained to select the pseudo-label.  
In~\cite{wang2018towards}, a box detected on an unlabeled image is pasted onto a labeled image, and the localization consistency estimation is performed onto the pasted label image. As the image itself is modified, a very thorough detection process is required in~\cite{wang2018towards}. In our method, only the lightweight detection head is processed. 
STAC~\cite{sohn2020simple} proposes to use a weak data augmentation for model training and a strong data augmentation is used for performing pseudo-label. However, like other pseudo-label methods~\cite{radosavovic2018data, zoph2020rethinking, li2020improving, sohn2020simple, wang2018towards}, it also follows the multi-stage training scheme. In contrast, our method is an end-to-end pseudo-labeling framework, which avoids the complicated training process and also achieves better performance.

\paragraph{Object Detection}
Object detection focuses on designing efficient and accurate detection framework. There are two mainstreams: single-stage object detectors~\cite{liu2016ssd,redmon2016you,tian2019fcos} and two-stage object detectors~\cite{girshick2015fast, ren2015faster, lin2017feature, yang2019reppoints, yang2019dense}. The main difference between the two types of methods is whether to use a cascade to filter a large number  of object candidates (proposals). In theory, our method is compatible with both types of methods. However, to allow a fair comparison with previous works~\cite{tang2021proposal,sohn2020simple} on semi-supervised object detection, we use Faster R-CNN~\cite{ren2015faster} as our default detection framework to illustrate our method.

\section{Methodology}
Figure.~\ref{fig:pipeline} illustrates an overview of our end-to-end training framework. There are two models, a student model and a teacher model. The student model is learned by both the detection losses on the labeled images and on the unlabeled images using pseudo boxes. The unlabeled images have two sets of pseudo boxes, which are used to drive the training of the classification branch and the regression branch, respectively. The teacher model is an exponential moving average (EMA) of the student model. Within this end-to-end framework, there are two crucial designs: \emph{soft teacher} and \emph{box jittering}.

\subsection{End-to-End Pseudo-Labeling Framework}
We first introduce the end-to-end framework for pseudo-label based semi-supervised object detection. Our approach follows the teacher-student training scheme. In each training iteration, labeled images and unlabeled images are randomly sampled according to a data sampling ratio $s_r$ to form a training data batch. The teacher model is performed to generate the pseudo boxes on unlabeled images and the student model is trained on both labeled images with the ground-truth and unlabeled images with the pseudo boxes as the ground-truth. Thus, the overall loss is defined as the weighted sum of supervised loss and unsupervised loss:
\begin{equation}
    \mathcal{L} = \mathcal{L}_s + \alpha \mathcal{L}_u, 
\end{equation}
where $\mathcal{L}_s$ and $\mathcal{L}_u$ denote supervised loss of labeled images and unsupervised loss of unlabeled images respectively, $\alpha$ controls contribution of unsupervised loss. Both of them are normalized by the respective number of images in the training data batch:

\begin{equation}
\mathcal{L}_s=\frac{1}{N_l}\sum_{i=1}^{N_l}(\mathcal{L}_{\text{cls}}(I_l^i) + \mathcal{L}_{\text{reg}}(I_l^i)),
\end{equation}
\begin{equation}\label{equ::unlabled_loss}
\mathcal{L}_u=\frac{1}{N_u}\sum_{i=1}^{N_u}(\mathcal{L}_{\text{cls}}(I_u^i) + \mathcal{L}_{\text{reg}}(I_u^i)),
\end{equation}
where $I_l^i$ indicates the $i$-th labeled image, $I_u^i$ indicates the $i$-th unlabeled image, $\mathcal{L}_{\text{cls}}$ is the classification loss, $\mathcal{L}_{\text{reg}}$ is the box regression loss, $N_l$ and $N_u$ denote the number of labeled images and unlabeled images, respectively.

At the beginning of training, both the teacher model and student model are randomly initialized. As the training progresses, the teacher model is continuously updated by the student model, and we follow the common practices~\cite{tarvainen2017mean, sohn2020fixmatch} that the teacher model is updated by exponential moving average (EMA) strategy.

In contrast to taking a simple probability distribution as the pseudo-label in image classification, creating pseudo-label for object detection is more complicated since an image usually contains multiple objects and the annotation of objects consists of location and category. Given an unlabeled image, the teacher model is used to detect objects and thousands of box candidates are predicted. The non-maximum suppression (NMS) is then performed to eliminate redundancy. Although most redundant boxes are removed, there are still some non-foreground candidates left. Therefore, only candidates with the foreground score\footnote{The foreground score is defined as the maximum probability of all non-background categories.} higher than a threshold are retained as the pseudo boxes. 

In order to generate high-quality pseudo boxes and to facilitate the training of the student model, we draw on FixMatch~\cite{sohn2020fixmatch} which is the latest advancement in semi-supervised image classification task. Strong augmentation is applied for detection training of the student model and weak augmentation is used for pseudo-labeling of the teacher model.

In theory, our framework is applicable to mainstream object detectors, including single-stage object detectors~\cite{lin2017focal,liu2016ssd,redmon2016you,tian2019fcos} and two-stage object detectors~\cite{ren2015faster,hu2018relation,chen2020reppoints,yang2019dense,yang2019reppoints}.
To allow a fair comparison with previous methods, we use Faster R-CNN~\cite{ren2015faster} as our default detection framework to illustrate our method.

\begin{figure*}
    \centering
    \includegraphics[width=1.0\linewidth]{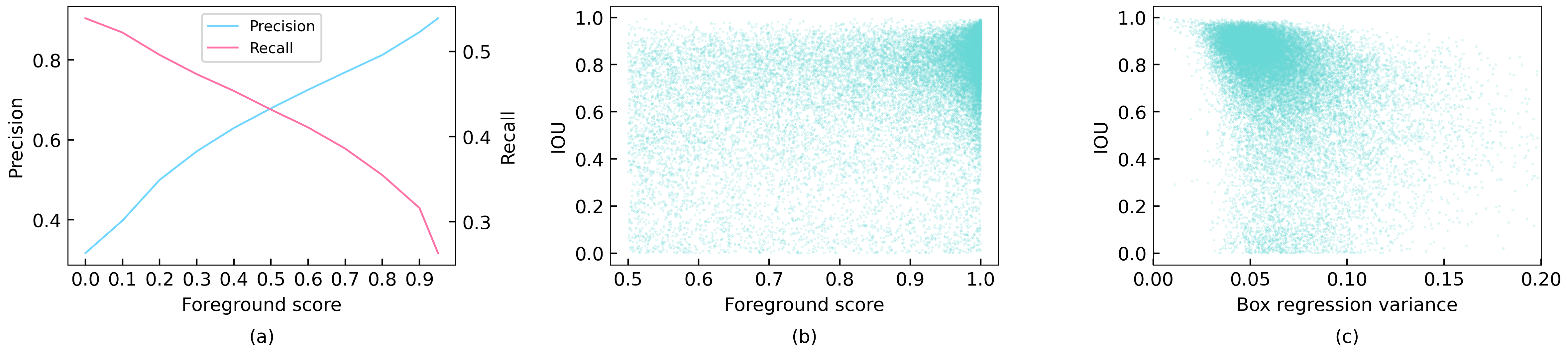}
    \caption{We randomly sampled 10k unlabeled training images from \texttt{train2017} to draw figures based on the model trained with $10\%$ labeled images. (a) precision and recall of foreground under different foreground score thresholds. (b) the correlation between the IoU with ground-truth and box foreground score. (c) the correlation between the IoU with ground-truth and box regression variance. Each point in (b) and (c) represents a box candidate. }
    \label{fig:analysis}
\end{figure*}

\subsection{Soft Teacher}
\label{sec::background_classification_reliability}
The performance of the detector depends on the quality of the pseudo-label. In practice, we find that using a higher threshold on foreground score to filter out most of the student-generated box candidates with low-confidence can achieve better results than using a lower threshold.
As shown in Table.~\ref{tab::ablation_threshold}, the best performance is achieved when the threshold is set to 0.9. However, while the strict criteria (higher threshold) leads to higher foreground precision, the recall of the retained box candidates also falls off quickly. As shown in Figure.~\ref{fig:analysis} (a), when the foreground threshold is set to 0.9, the recall is low, as $33\%$, while the precision reaches $89\%$. In this case, if we use IoU between student generated box candidates and teacher-generated pseudo boxes to assign foreground and background labels, as a general object detection framework does when real box annotations are provided, some foreground box candidates will be mistakenly assigned as negatives, which may hinder the training and harm the performance.

To alleviate this issue, we propose a \emph{soft teacher} approach which leverages richer information from the teacher model, thanks to the flexibility of the end-to-end framework. Specifically, we assess the reliability of each student-generated box candidate to be a real background, which is then used to weigh its background classification loss. Given two box sets $\{b_i^{\text{fg}}\}$ and $\{b_i^{\text{bg}}\}$, with $\{b_i^{\text{fg}}\}$ denoting boxes assigned as foreground and $\{b_i^{\text{bg}}\}$ denoting the boxes assigned as background, the classification loss of an unlabeled image with the reliable weighting is defined as:

\begin{equation}\label{equ::unlabled_cls}
\begin{aligned}
    \mathcal{L}_u^{\text{cls}} = \frac{1}{N_b^{\text{fg}}}\sum_{i=1}^{N_b^{\text{fg}}} l_{\text{cls}}(b_i^{\text{fg}}, \mathcal{G_{\text{cls}}}) + \sum_{j=1}^{N_b^{\text{bg}}} w_j l_{\text{cls}}(b_j^{\text{bg}}, \mathcal{G_{\text{cls}}}),
\end{aligned}
\end{equation}

\begin{equation}
    w_j = \frac{r_j}{\sum_{k=1}^{N_b^{\text{bg}}} r_k},
\end{equation}
where $\mathcal{G_{\text{cls}}}$ denotes the set of (teacher-generated) pseudo boxes used for classification, $l_{\text{cls}}$ is the box classification loss, $r_j$ is the reliability score for j-th background box candidate, $N_b^{\text{fg}}$ and $N_b^{\text{bg}}$ are the number of box candidates of the box set $\{b_i^{\text{fg}}\}$ and $\{b_i^{\text{bg}}\}$, respectively. 

Estimating the reliability score $r$ is challenging. We find empirically that the background score produced by the teacher model with weak augmented image can well serve as a proxy indicator of $r$ and is easily obtained in our end-to-end training framework. Specifically, given a student-generated box candidate, its background score can be obtained simply by using the teacher (BG-T) to process the box through its detection head. It is worth noting that this approach, unlike the widely used hard negative mining approaches, e.g., OHEM~\cite{shrivastava2016training} or Focal Loss~\cite{lin2017focal}, is more like a “simple” negative mining.
For comparison, we also examine several other indicators:

\begin{itemize}
\item \emph{Background score of student model (BG-S)}: Another natural way to generate the background score is to use the prediction of student model directly.

\item \emph{Prediction difference (Pred-Diff)}: The prediction difference between the student model and teacher model is also a possible indicator. In our approach, we simply use the difference between the background scores of the two models to define the reliability score:
\begin{equation}
    r = 1 - |p_{\text{S}}^{\text{bg}}(b) - p_{\text{T}}^{\text{bg}}(b)|,
\end{equation}
where $p_{\text{S}}^{\text{bg}}$ and $p_{\text{T}}^{\text{bg}}$ are the predicted probability of the background class of the student and the teacher model, respectively.  

\item \emph{Intersection-over-Union}: The IoU between ground-truths and box candidates is a commonly used criterion for foreground/background assignment. There are two different hypotheses about how to use IoU to measure whether a box candidate belongs to the background. In the first hypothesis, if the IoU between a box candidate and a ground-truth box is less than a threshold (e.g., 0.5), a larger IoU indicates the box candidate has greater probability of being background. This can be viewed as an IoU-based hard negative mining which is adopted by Fast R-CNN~\cite{girshick2015fast} and Faster R-CNN~\cite{ren2015faster} in the early implementation. In contrast, the other hypothesis suggests that box candidates with a smaller IoU with ground-truths are more likely to be backgrounds. In our experiments, we validate both hypotheses and name them as \emph{IoU} and \emph{Reverse-IoU}.
\end{itemize}

\subsection{Box Jittering}
As shown in Figure.~\ref{fig:analysis} (b), the localization accuracy and the foreground score of the box candidates do not show a strong positive correlation, which means that the boxes with high foreground score may not provide accurate localization information. This indicates that the selection of the teacher-generated pseudo boxes according to the foreground score is not suitable for box regression, and a better criterion is needed.

We introduce an intuitive approach to estimate the localization reliability of a candidate pseudo box by measuring the consistency of its regression prediction. Specifically, given a teacher-generated pseudo box candidate $b_i$, we sample a jittered box around $b_i$ and feed the jittered box into the teacher model to obtain the refined box $\hat{b}_{i}$, which is formulated as follows:
\begin{equation}
    \hat{b}_{i} = \text{refine}(\text{jitter}(b_i)).
\end{equation}
The above procedure is repeated several times to collect a set of $N_{\text{jitter}}$ refined jittered boxes $\{\hat{b}_{i,j}\}$, and we define the localization reliability as the box regression variance:
\begin{equation}
    \bar{\sigma_{i}} = \frac{1}{4} \sum_{k=1}^{4}\hat{\sigma_k},
\end{equation}
\begin{equation}
    \hat{\sigma_k} = \frac{\sigma_k}{0.5(h(b_i) + w(b_i))},
\end{equation}
where $\sigma_k$ is the standard derivation of the k-th coordinate of the refined jittered boxes set $\{\hat{b}_{i,j}\}$, $\hat{\sigma_k}$ is the normalized $\sigma_k$, $h(b_i)$ and $w(b_i)$ represent the height and width of box candidate $b_i$, respectively.

A smaller box regression variance indicates a higher localization reliability. However, computing the box regression variances of all pseudo box candidates is unbearable during training. Therefore, in practice, we only calculate the reliability for the boxes with a foreground score greater than 0.5. In this way, the number of boxes that need to be estimated is reduced from an average of hundreds to around 17 per image and thus the computation cost is almost negligible.

In Figure.~\ref{fig:analysis} (c), we illustrate the correlation between the localization accuracy and our box regression variance. Compared with the foreground score, the box regression variance can better measure the localization accuracy. This motivates us to select box candidates whose box regression variance is smaller than a threshold as pseudo-label to train the box regression branch on unlabeled images. Given the pseudo boxes $\mathcal{G}_{\text{reg}}$ for training the box regression on unlabeled data, the regression loss is formulated as:
\begin{equation}\label{equ::unlabled_reg}
\begin{aligned}
    \mathcal{L}_u^{\text{reg}} = \frac{1}{N_b^{\text{fg}}}\sum_{i=1}^{N_b^{\text{fg}}} l_{\text{reg}}(b_i^{\text{fg}}, \mathcal{G}_{\text{reg}}),
\end{aligned}
\end{equation}
where $b_i^{\text{fg}}$ is i-th box assigned as foreground, $N_b^{\text{fg}}$ is the total number of foreground box, $l_{\text{reg}}$ is the box regression loss. Therefore, by substituting  Equ.~\ref{equ::unlabled_cls} and Equ.~\ref{equ::unlabled_reg} into Equ.~\ref{equ::unlabled_loss}, the loss of unlabeled images is:
\begin{equation}
\begin{aligned}
    \mathcal{L}_u=\frac{1}{N_u}\sum_{i=1}^{N_u}(\mathcal{L}_u^{\text{cls}}(I_u^i, \mathcal{G}_{\text{cls}}^i) + \mathcal{L}_u^{\text{reg}}(I_u^i, \mathcal{G}_{\text{reg}}^i)).
\end{aligned}
\end{equation}
Here we use the pseudo boxes $\mathcal{G}_{\text{cls}}$ and $\mathcal{G}_{\text{reg}}$ as the inputs of the loss to highlight the fact that the pseudo boxes used in classification and box regression are different in our approach.

\begin{table*}
\begin{center}
\begin{tabular}{c|c|c|c}
\hline
 Augmentation & Labeled image training & Unlabeled image training & Pseudo-label generation\\
\hline
Scale jitter &short edge $\in(0.5,1.5)$ &short edge $\in(0.5,1.5)$ &short edge $\in(0.5,1.5)$ \\
Solarize jitter & p=0.25, ratio $\in(0,1)$ & p=0.25, ratio $\in(0,1)$  &- \\
Brightness jitter & p=0.25, ratio $\in(0,1)$ & p=0.25, ratio $\in(0,1)$ & - \\
Constrast jitter & p=0.25, ratio $\in (0,1)$ & p=0.25, ratio $\in (0,1)$ & - \\
Sharpness jitter & p=0.25, ratio $\in (0,1)$ & p=0.25, ratio $\in (0,1)$ & - \\
Translation &-&p=0.3, translation ratio $\in (0,0.1)$&-\\
Rotate &-&p=0.3, angle $\in (0,30^{\circ} )$&-\\
Shift&-&p=0.3, angle $\in (0,30^{\circ} )$&-\\
Cutout &num $\in(1,5)$, ratio $\in(0.05,0.2)$ &num $\in(1,5)$, ratio $\in(0.05,0.2)$ & -  \\
\hline
\end{tabular}
\end{center}
\caption{The summary of the data augmentation used in our approach. We follow the practice of STAC~\cite{sohn2020simple} and FixMatch~\cite{sohn2020fixmatch} to provide different data augmentation for pseudo-label generation, labeled image training and unlabeled image training. ``-'' indicates the augmentation is not used.}
\label{tab::data_aug}
\end{table*}

\begin{table*}
\begin{center}
\begin{tabular}{c|c|c|c}
\hline
Method & 1\% & 5\% & 10\% \\
\hline
Supervised baseline (Ours) & $10.0\pm{0.26}$ & $20.92\pm{0.15}$ & $26.94\pm{0.111}$\\
Supervised baseline (STAC)~\cite{sohn2020simple} & $9.83\pm{0.23}$ & $21.18\pm{0.20}$ & $26.18\pm{0.12}$ \\
STAC~\cite{sohn2020simple} & $13.97\pm{0.35}$ & $24.38\pm{0.12}$ & $28.64\pm{0.21}$  \\
Ours & \textbf{20.46}$\pm{0.39}$ & \textbf{30.74}$\pm{0.08}$ & \textbf{34.04}$\pm{0.14}$ \\
\hline
\end{tabular}
\end{center}
\caption{System level comparison with STAC on \texttt{val2017} under the \textbf{Partially Labeled Data} setting. All the results are the average of all 5 folds. For benchmarking, we also compare the supervised benchmark performance between our method and STAC, and their performance is similar.}
\label{tab::system_comparision_coco}
\end{table*}

\section{Experiments}
\subsection{Dataset and Evaluation Protocol}
We validate our method on the MS-COCO benchmark~\cite{lin2014microsoft}. Two training datasets are provided, the \texttt{train2017} set contains 118k labeled images and the \texttt{unlabeled2017} set contains 123k unlabeled images. In addition, the \texttt{val2017} set with 5k images is also provided for validation. In previous methods~\cite{sohn2020simple, tang2021proposal,jeong2019consistency}, there are two settings for validating the performance:

\noindent \textbf{Partially Labeled Data}: STAC~\cite{sohn2020simple} first introduced this setting. 1\%, 5\% and 10\% images of \texttt{train2017} set are sampled as the labeled training data, and the remaining unsampled images of \texttt{train2017} are used as the unlabeled data. For each protocol, STAC provides 5 different data folds and the final performance is the average of all 5 folds.

\noindent \textbf{Fully Labeled Data}: In this setting, the entire \texttt{train2017} is used as the labeled data and \texttt{unlabeled2017} is used as the additional unlabeled data. This setting is more challenging. Its goal is to use the additional unlabeled data to improve a well-trained detector on large-scale labeled data. 

We evaluate our method on both  settings and follow the convention to report the performance on \texttt{val2017} with the standard mean average precision (mAP) as the evaluation metrics.

\begin{table*}[!h]
\begin{center}
\begin{tabular}{c|c|c}
\hline
Method & Extra dataset & mAP \\
\hline
Proposal learning~\cite{tang2021proposal} & unlabeled2017 & $37.4\xrightarrow{\text{+1.0}}38.4$\\
STAC~\cite{sohn2020simple} & unlabeled2017 & $39.5\xrightarrow{\text{-0.3}}39.2$ \\
Self-training~\cite{zoph2020rethinking} & ImageNet+OpenImages &  $41.1\xrightarrow{\text{+0.8}}41.9$ \\
Ours & unlabeled2017 & $40.9\xrightarrow{\textbf{+3.6}}44.5$ \\
\hline
\end{tabular}
\end{center}
\caption{Comparison with other state-of-the-arts under the setting of using all data of \texttt{train2017} set. Particularly, Self-training uses ImageNet (1.2M images) and OpenImages (1.7M images) as additional unlabeled images, which is 20$\times$ larger than \texttt{unlabeled2017} (123k images).}
\label{tab::system_full_coco_others}
\end{table*}

\begin{table*}
\small
\begin{center}
\begin{tabular}{c|c|c|c|c}
\hline
detector & backbone & method & $\text{mAP}^{\text{det}}$ & $\text{mAP}^{\text{mask}}$  \\
\hline
\multirow{2}{*}{Faster R-CNN} & \multirow{2}{*}{ResNet-50} & supervised & 40.9 & - \\
& & ours & 44.5(\textbf{+3.6}) & - \\
\hline
\multirow{2}{*}{Faster R-CNN} & \multirow{2}{*}{ResNet-101} & supervised & 43.8 & - \\
& & ours & 46.8(\textbf{+3.0}) & - \\
\hline
\multirow{2}{*}{HTC++} & \multirow{2}{*}{Swin-L} & supervised & 57.1 & 49.6\\
& & ours & 59.1(\textbf{+2.0}) & 51.0(\textbf{+1.4}) \\
\hline
\multirow{2}{*}{HTC++(multi-scale)} & \multirow{2}{*}{Swin-L} & supervised & 58.2 & 50.5 \\
& & ours & 59.9(\textbf{+1.7}) & 51.9(\textbf{+1.4})\\
\end{tabular}
\end{center}
\caption{Compared with various supervised trained detectors on \texttt{val2017}. The entire \texttt{train2017} is used as the labeled images, and the \texttt{unlabeled2017} is used as the additional unlabeled images. }
\label{tab::system_full_coco_more_results}
\end{table*}

\subsection{Implementation Details}
We use the Faster R-CNN~\cite{ren2015faster} equipped with FPN~\cite{lin2017feature} (Feature Pyramid Network) as our default detection framework to evaluate the effectiveness of our method, and an ImageNet pre-trained ResNet-50~\cite{he2016deep} is adopted as the backbone. Our implementation and hyper-parameters are based on MMDetection~\cite{chen2019mmdetection}. Anchors with 5 scales and 3 aspect ratios are used. 2k and 1k region proposals are generated with a non-maximum suppression threshold of 0.7 for training and inference. In each training step, 512 proposals are sampled from 2k proposals as the box candidates to train RCNN. Since the amount of training data of \textbf{Partially Labeled Data} setting and \textbf{Full Labeled Data} setting has large differences, the training parameters under the two settings are slightly different.

\noindent\textbf{Partially Labeled Data}: The model is trained for 180k iterations on 8 GPUs with 5 image per GPU. With SGD training, the learning rate is initialized to 0.01 and is divided by 10 at 110k iteration and 160k iteration. The weight decay and the momentum are set to 0.0001 and 0.9, respectively. The foreground threshold is set to 0.9 and the data sampling ratio $s_r$ is set to 0.2 and gradually decreases to 0 over the last 10k iterations.

\noindent\textbf{Fully Labeled Data}: The model is trained for 720k iterations on 8 GPUs with 8 image per GPU. In SGD training, the learning rate is initialized to 0.01 and is divided by 10 at 480k iteration and 680k iteration. The weight decay and the momentum are set to 0.0001 and 0.9, respectively. The foreground threshold is set to 0.9 and the data sampling ratio $s_r$ is set to 0.5 and gradually decreases to 0 in the last 20k iterations. 

For estimating the box localization reliability, we set $N_{\text{jitter}}$ as 10, and threshold is set as 0.02 to select the pseudo-labels for box regression. The jittered boxes are randomly sampled by adding the offsets on four coordinates, and the offsets are uniformly sampled from [-6\%, 6\%] of the height or width of the pseudo box candidates. In addition, we follow STAC and FixMatch to use different data augmentation for pseudo-label generation, labeled image training and unlabeled image training. The details are summarized in Table~.\ref{tab::data_aug}.

\begin{table}
\begin{center}
\begin{tabular}{c|c|c|c}
\hline
Method & mAP & mAP@0.5 & mAP@0.75 \\
\hline
Supervised & 27.1 & 44.6 & 28.6\\
Multi-Stage & 28.7 & 47.0 & 30.9 \\
E2E & 30.0 & 47.4 & 32.4 \\
E2E+EMA & \textbf{31.2} & \textbf{48.8} & \textbf{34.0}  \\
\hline
\end{tabular}
\end{center}
\caption{Multi-Stage \textit{vs.} End-to-End. The end-to-end (E2E) method outperforms the multi-stage framework. Updating the teacher network through the exponential moving average (EMA) strategy further improves the performance.}
\label{tab::ablation_e2e}
\end{table}

\begin{table*}
\begin{center}
\begin{tabular}{c@{~}|@{~}c@{~}|@{~}c@{~}|@{~}c|c}
\hline
Soft teacher & Box jittering & mAP & mAP@0.5 & mAP@0.75 \\
\hline
& & 31.2 & 48.8 & 34.0 \\
\checkmark & &  33.6 & \textbf{52.9} & 36.6 \\
\checkmark & \checkmark & \textbf{34.2} & 52.6 & \textbf{37.3} \\
\hline
\end{tabular}
\end{center}
\caption{We study the effects of \emph{soft teacher} and \emph{box jittering} techniques. }
\label{tab::ablation_component}
\end{table*}

\begin{table}
\begin{center}
\begin{tabular}{c|c|c|c}
\hline
 Indicator & mAP & mAP@0.5 & mAP@0.75 \\
\hline
w/o weight & 31.2 & 48.8 & 34.0 \\
IoU & 31.7 & 51.4 & 34.2 \\
Reverse-IoU & 31.6 & 49.5 & 34.1 \\
Pred-Diff & 32.3 & 51.0 & 34.6 \\
BG-S & 25.9 & 44.4 & 27.0 \\
BG-T & \textbf{33.6} & \textbf{52.9} & \textbf{36.6} \\
\hline
\end{tabular}
\end{center}
\caption{Comparison of different indicators in soft teacher. }
\label{tab::ablation_different_weighting}
\end{table}

\subsection{System Comparison}
\begin{figure}
    \centering
    \includegraphics[width=1.0\linewidth]{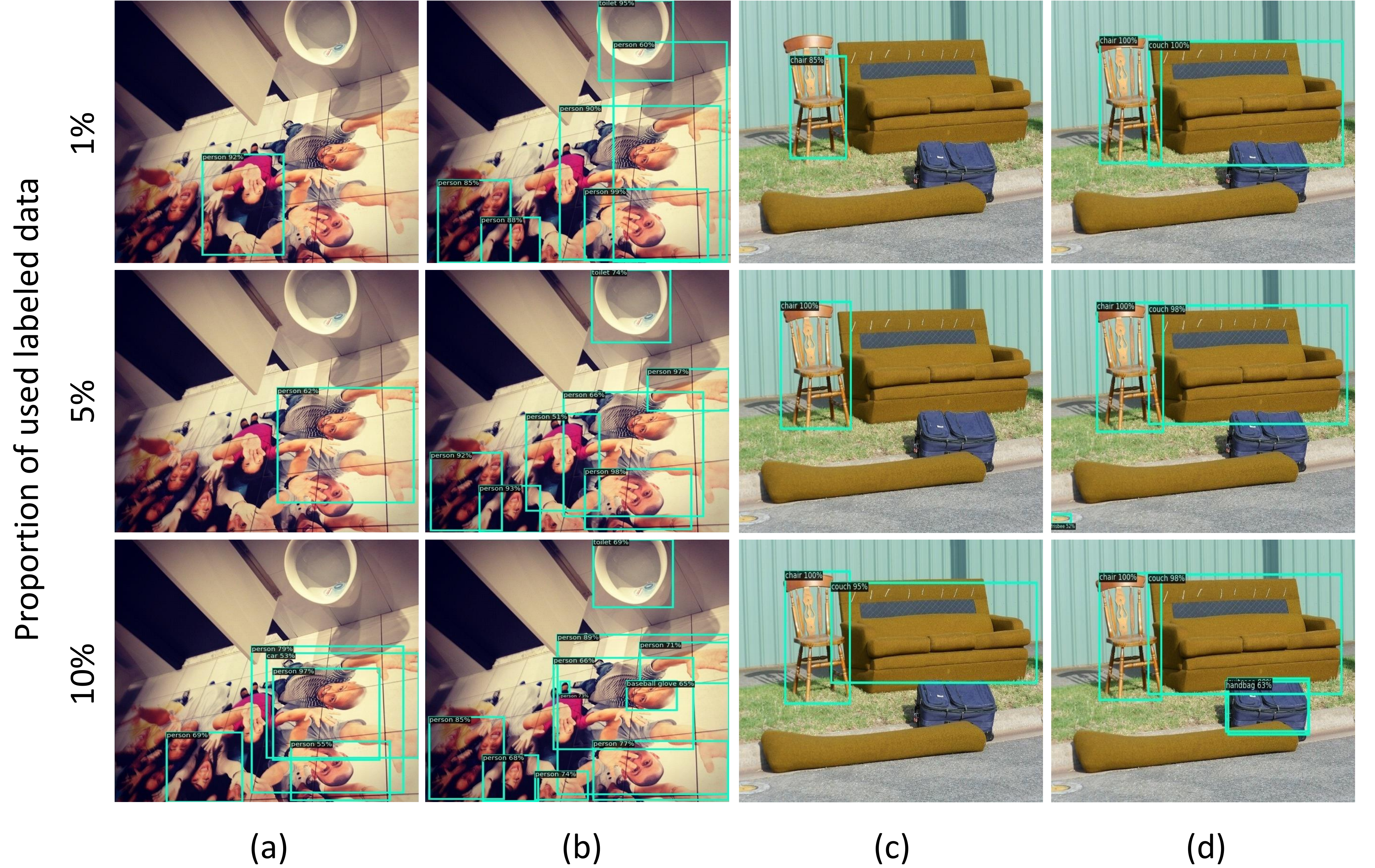}
    \caption{The qualitative results of our method. (a), (c) are the results of the supervised baseline. (b), (d) are the results of our method. }
    \label{fig:detection_results}
\end{figure}

In this section, we compare our method with previous state-of-the-arts on MS-COCO. We first evaluate on the \textbf{Partially Labeled Data} setting and compare our method with STAC. For benchmarking, we compare the supervised baseline of our method with the results reported in STAC and find they perform similarly, the results are shown in Table.~\ref{tab::system_comparision_coco}. In this case, we further compare our method with STAC at the system level, and our method shows a significant performance improvement in different protocols. Specifically, our method outperforms the STAC by \textbf{6.5} points, \textbf{6.4} points and \textbf{5.4} points when there are $1\%$, $5\%$, and $10\%$ labeled data, respectively. The qualitative results of our method compared with supervised baseline are shown in Figure.~\ref{fig:detection_results}.

Then we compare our method with other state-of-the-art methods in \textbf{Fully Labeled Data} setting. Since the reported performance of supervised baseline varies in different works, we report the results of the comparison methods and their baseline at the same time. The results are shown in Table.~\ref{tab::system_full_coco_others}.

We first compare with the Proposal Learning~\cite{tang2021proposal} and STAC~\cite{sohn2020simple} which also use \texttt{unlabeled2017} as additional unlabeled data. Because of the better hyper-parameters and more adequate training, our supervised baseline achieved better performance than other methods. Under the stronger baseline, our method still shows a greater performance gain (+3.6 points) than Proposal Learning (+1.0 points) and STAC (-0.3 points). Self-training~\cite{zoph2020rethinking} uses ImageNet (1.2M images) and OpenImages (1.7M images) as the additional unlabeled data, which is 20$\times$ larger than the \texttt{unlabeled2017} (123k images) that we use. With similar baseline performance, our method also shows better result with less unlabeled data.

 In addition, we further evaluate our method on other stronger detectors, and the results evaluated on \texttt{val2017} set are shown in Table.~\ref{tab::system_full_coco_more_results}. Our method consistently improves the performance of different detectors by a notable margin. Even in the state-of-the-art detector HTC++ with Swin-L backbone, we still show 1.8 improvement on detection AP and 1.4 improvement on mask AP. Moreover, we also report the results on \texttt{test-dev2017} set. As shown in Tabel.~\ref{tab::teaser_table}, our method improves the HTC++ with Swin-L backbone by 1.5 mAP on detection, which is the first work to surpass 60 mAP on COCO object detection benchmark.
 
\subsection{Ablation Studies}
 In this section, we validate our key designs. If not specified, all the ablation experiments are performed on the single data fold provided by \cite{sohn2020simple} with 10\% labeled images from \texttt{train2017} set.

\paragraph{Multi-Stage \textit{vs.} End-to-End.} 
We compare our end-to-end method with the multi-stage framework as shown in Table~\ref{tab::ablation_e2e}. By simply switching from the multi-stage framework to our end-to-end framework, performance is increased by 1.3 points. By updating the teacher model with the student model through the exponential moving average (EMA) strategy, our method further achieves 31.2 mAP. 

\paragraph{Effects of Soft Teacher and Box Jittering.}
We ablate the effects of soft teacher and box jittering. The results are shown in Table.~\ref{tab::ablation_component}. Based on our end-to-end model equipped with EMA (E2E+EMA), integrating the soft teacher improves the performance by 2.4 points. Further applying the box jittering, the performance reaches 34.2 mAP, which is 3 points better than E2E+EMA.

\paragraph{Different Indicators in Soft Teacher.}
In Section.~\ref{sec::background_classification_reliability}, several different indicators are explored for reliability estimation. Here, we evaluate the different indicators and the results are shown in Table.~\ref{tab::ablation_different_weighting}. The background score predicted by the teacher model achieves the best performance. Simply switching the model from teacher to student will make the performance worse. In addition, the improvement of IoU and Revearse-IoU is negligible compared with BG-T. These results prove the necessity of leveraging the teacher model.

\begin{table}
\begin{center}
\begin{tabular}{c|c|c|c}
\hline
Threshold & mAP & mAP@0.5 & mAP@0.75 \\
\hline
0.70 & 29.9 & 48.6 & 32.1\\
0.80 & 33.2 & 52.8 & 35.9\\
0.90 & \textbf{33.6} & \textbf{52.9} & \textbf{36.6}\\
0.95 & 32.1 & 50.6 & 34.7\\
\hline
\end{tabular}
\end{center}
\caption{Ablation study on the effects of different foreground thresholds. }
\label{tab::ablation_threshold}
\end{table}

\begin{table}
\begin{center}
\begin{tabular}{c|c|c|c}
\hline
Threshold & mAP & mAP@0.5 & mAP@0.75 \\
\hline
0.04 & 33.8 & 52.3 & 36.7\\
0.03 & 34.0 & 52.5 & 36.9\\
0.02 & \textbf{34.2} & \textbf{52.6} & \textbf{37.3}\\
0.01 & 32.9 & 52.2 & 35.8\\
\hline
\end{tabular}
\end{center}
\caption{Ablation study on the effects of different thresholds for selecting pseudo boxes for box regression according to box regression variance. }
\label{tab::ablation_loc_reliable}
\end{table}

\begin{table}[!h]
\begin{center}
\begin{tabular}{c|c|c|c}
\hline
$N_{\text{jitter}}$ & mAP & mAP@0.5 & mAP@0.75 \\
\hline
5 & 34.0 & 52.3 &37.0 \\
10 & \textbf{34.2} & \textbf{52.6} & 37.3\\
15 & \textbf{34.2} & 52.5 & \textbf{37.4}\\
\hline
\end{tabular}
\end{center}
\caption{Ablation study on the effects of different number of jittered boxes used to estimate the box regression variance. }
\label{tab::ablation_box_num}
\end{table}

\paragraph{Effects of other hyper-parameters.}
We study the effects of hyper-parameters used in our method. Table.~\ref{tab::ablation_threshold} studies the effects of different foreground score thresholds. The best performance is achieved when the threshold is set to 0.9, and lower or higher thresholds will cause significant performance degradation. In Table.~\ref{tab::ablation_loc_reliable}, we study the box regression variance threshold. The best performance is achieved when the threshold is set to 0.02. In Table.~\ref{tab::ablation_box_num}, we study the effects of different number of jittered boxes, and the performance is saturated when $N_{\text{jitter}}$ is set to 10.

\section{Conclusion}
In this paper, we proposed an end-to-end training framework for semi-supervised object detection, which discards the complicated multi-stage schema adopted by previous approaches. Our method simultaneously improves the detector and pseudo labels by leveraging a student model for detection training, and a teacher model which is continuously updated by the student model through the exponential moving average strategy for online pseudo-labeling. Within the end-to-end training, we present two simple techniques named soft teacher and box jittering to facilitate the efficient leverage of the teacher model. The proposed framework outperforms the state-of-the-art methods by a large margin on MS-COCO benchmark in both partially labeled data and fully labeled data settings.

\section{Acknowledgement}
We would like to thank Yue Cao for his valuable suggestions and discussions; Yutong Lin and Yixuan Wei for help on Swin Transformer experiments.

{\small
\bibliographystyle{ieee_fullname}
\bibliography{egbib}
}

\end{document}